# Research on Stable Obstacle Avoidance Control Strategy for Tracked Intelligent Transportation Vehicles in Non-structural Environment Based on Deep Learning

Yitian Wang, Jun Lin, Liu Zhang, Tianhao Wang, Hao Xu, Guanyu Zhang, Yang Liu

*Abstract*—Existing intelligent driving technology often has a problem in balancing smooth driving and fast obstacle avoidance, especially when the vehicle is in a non-structural environment, and is prone to instability in emergency situations. Therefore, this study proposed an autonomous obstacle avoidance control strategy that can effectively guarantee vehicle stability based on Attention-long short-term memory (Attention-LSTM) deep learning model with the idea of humanoid driving. First, we designed the autonomous obstacle avoidance control rules to guarantee the safety of unmanned vehicles. Second, we improved the autonomous obstacle avoidance control strategy combined with the stability analysis of special vehicles. Third, we constructed a deep learning obstacle avoidance control through experiments, and the average relative error of this system was 15%. Finally, the stability and accuracy of this control strategy were verified numerically and experimentally. The method proposed in this study can ensure that the unmanned vehicle can successfully avoid the obstacles while driving smoothly.

## I. INTRODUCTION

With the accelerating process of vehicle intelligentization, humanoid autonomous obstacle avoidance driving, which is an emerging technology in the field of unmanned driving, has attracted more and more attention from scholars [1]-[4]. Among them, tracked intelligent transport vehicles are developed with more and more important functions and tasks, making them important equipment in special transport operations. However, they are often accompanied by safety hazards, such as collision and rollover during obstacle avoidance driving, since the working scenes of tracked intelligent transportation vehicles are mostly non-structural and complex environments, making it difficult to maintain the stability of the vehicle [5] [6]. Therefore, it is important to study how to improve the intelligent driving capability of tracked intelligent transport vehicles in non-structural environments.

For many years, tracked intelligent transportation vehicles have been the focus of various researchers. They have conducted research on autonomous obstacle avoidance methods for vehicles [7]-[12]. Hu et al. proposed a path planning method based on the off-road environment [13]. Schaub et al. introduced a two-stage clustering of the optical flow and proposed how those clusters can be used for obstacle evasion. Using this method, a simulation test was performed on the test site, and static and dynamic obstacles were successfully avoided [14]. Gim et al. proposed a new continuous curvature path generation method for poor comfort and low steering efficiency during vehicle obstacle avoidance lane change maneuvers [15]. Gla B. et al. proposed a risk assessment-based collision avoidance decision algorithm for multi-scene autonomous vehicles [16]. Li et al. proposed a method combining local trajectory planning and tracking control to solve the problem of vehicle lane changing and obstacle avoidance [17].

Artificial intelligence has promising applications in various areas, such as image recognition, histological modeling, and assisted driving, due to the significant increase in computer power. Gordon et al. applied artificial intelligence to the field of optical and photonic systems [18]. Andrew W. et al. can further understand protein structures, functions, and obstacles by establishing a neural network model to predict the structure of the protein [19]. Ozturk et al. proposed an automatic chest image detection method based on the combination of the artificial intelligence technology and radiological imaging technology for the screening of patients with COVID-2019 [20]. Wang et al. used deep learning technology to predict the mutation status of EGFR in lung adenocarcinoma [21]. Xu et al. proposed a hierarchical active learning framework based on convolutional neural network (CNN) method to improve the accuracy of scene classification [22]. Wu. et al proposed a RUL prediction method using multi-sensor time series signals based on deep long short-term memory (DLSTM) [23].

In assisted driving, Shi et al. proposed a humanoid control method based on deep learning models to train driving data to predict braking decisions [24]. Chen et al. developed a vision-based deep Monte Carlo tree search (deep-MCTS) method for automatic driving control, which can predict driving behavior

This work was supported by the National Natural Science Foundation of China (Grant Nos. 41827803), Science and Technology Development Project of Changchun, China. (Grant Nos. 21ZY21). (*Corresponding author*: *Guanyu Zhang, Yang Liu*.)

Yitian Wang, Jun Lin, Liu Zhang, Tianhao Wang, Hao Xu, Guanyu Zhang, and Yang Liu are with the College of Instrumentation and Electrical engineering, Jilin University, Changchun 130026, China (e-mail: yitian18@mails.jlu.edu.cn; lin_jun@jlu.edu.cn; zhangliu@jlu.edu.cn; wangtianhao@jlu.edu.cn; xuh20@mails.jlu.edu.cn; zhangguanyu@jlu.edu.cn; liu_yang@jlu.edu.cn)

and help improve driving control stability and performance [25]. Sun et al. proposed a fuzzy algorithm-based driver driving behavior classification method and established a brain-based decision linear neural network to implement humanoid driving [26]. Liang et al. proposed a target detection algorithm based on machine vision and neural network to improve the safety of autonomous driving [27]. Liu et al. applied the neural network module to the three-dimensional target detection of automatic driving, which improved the prediction performance of the system [28]. Aladem et al. proposed a single-stream dual-task network that performed semantic segmentation and monocular depth without using multiple decoders, improving the system performance of self-driving vision perception [29]. Huang et al. proposed a prediction model based on LSTM neural network to capture real traffic flow features by introducing human driving memory [30]. Zhang et al. proposed a new MV-CNN model for training and recognition of driving behavior by learning sample data collected from in-vehicle sensors [31].

In summary, we constructed a humanoid driving model based on deep learning. First, a stable obstacle avoidance control strategy based on deep learning is proposed. Then, a training data set is established using a virtual prototype. Finally, the accuracy of the method is verified through simulation and experiment. The experimental results show that the method has good accuracy and stability.

## II. PROBLEM DESCRIPTION

Since the tracked intelligent transportation vehicle has a high center of gravity, and the field environment where the vehicle works is mostly located in mountainous areas with a slope greater than 5°, the vehicle will generate a large lateral acceleration when the tracked intelligent transportation vehicle is in emergency obstacle avoidance and turning state. Considering that there are a large number of trees, wild rocks and other obstacles on the hillside, the road is relatively rough, and the vehicle will continue to cross the undulating obstacles during the driving process, resulting in aggravated instability. The tipping situation is as follows:

a. Steering and tipping: When the vehicle is driving on a slope with a slope angle of β, the lateral acceleration of the vehicle is not zero due to the action of the ground gravity, the supporting force and the friction force of the slope, and the critical angular velocity of rollover changes, and the vehicle is prone to rollover, as shown in Fig. 1(a).

b. Longitudinal uphill and downhill tipping: Tracked intelligent transportation vehicles in the longitudinal uphill turning and obstacle avoidance process due to the increase in the slope angle makes the vehicle along the slope direction of the component force increases, making the vehicle produce along the slope face downward side tilt angle, resulting in the vehicle under the slope when the side tilt angle increases, as shown in Fig. 1(b). When the crawler type intelligent transportation vehicle is in the longitudinal downhill turning and obstacle avoidance process, the component force of the vehicle along the slope direction decrease due to the decrease of slope angle, making the vehicle produce the side inclination angle along the slope facing upward, resulting in the increase of the side inclination angle when the vehicle is under the slope, as shown in Fig. 1(b).

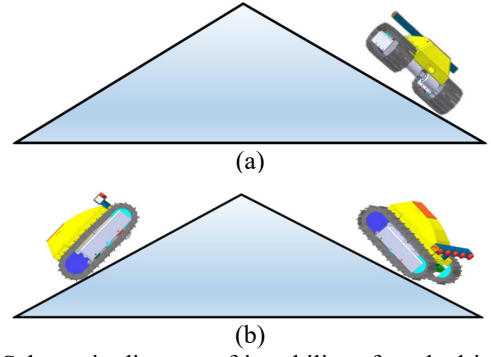

**Fig. 1.** Schematic diagram of instability of tracked intelligent transport vehicle. (a) Steering and tipping; (b) longitudinal uphill and downhill tipping.

As shown in Figure 2, it is difficult for unmanned vehicles to plan a stable and smooth obstacle avoidance path in the process of obstacle avoidance as professional drivers do due to the extremely complex working environment and operational requirements of tracked intelligent transportation vehicles, which has a great impact on the safety and stability of unmanned vehicles. Thus, we proposed an active control strategy to adjust the vehicle posture and improve the stability of vehicle obstacle avoidance.

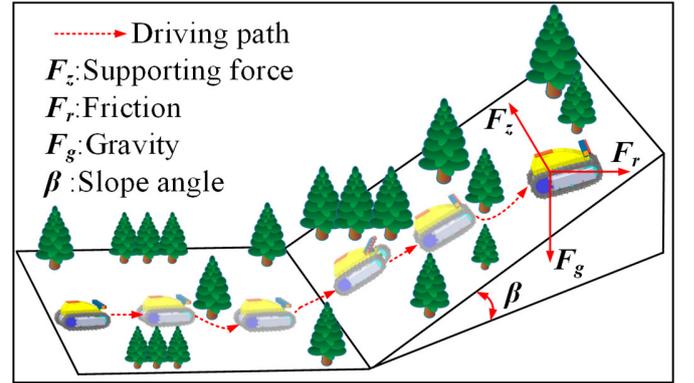

**Fig. 2.** Schematic diagram of the vehicle obstacle avoidance process based on a non-structural road.

## III. METHOD

First, we constructed an obstacle information detection system and proposed an obstacle classification strategy based on obstacle characteristics and spatial relative position relationship to classify the obstacles ahead into 15 situations. Second, we analyzed the stability index threshold of the vehicle under the limit state. Finally, we adjusted the vehicle driving control strategy under the current emergency based on the location and state information of the obstacles and vehicles, respectively, as the input of the deep learning model.

### A. Obstacle Avoidance Strategy

*1) Analysis of Spatial Relative Position Relationship of Detection Target*

Ultrasonic sensors use high-frequency sound waves to detect the position and distance of objects with strong directionality and can be used to obtain obstacle information of the surrounding environment. A proper sensor layout can effectively reduce the blind spot of obstacle perception. Therefore, in this study, we constructed an obstacle information detection system based on the ultrasonic sensor.

In engineering applications, it is necessary to ensure the successful obstacle avoidance of vehicles and the easy realization of the design of the decision planning. The tracked intelligent transport vehicles need to use the obstacle information detection system to autonomously classify field obstacles due to the complex field environment, harsh working conditions, and no apparent marking guidelines and traffic signals on non-structural roads. Therefore, we proposed an obstacle classification strategy based on obstacle characteristics and spatial relative position relationships. First, the ultrasonic sensor in front of the vehicle head needs to be used to detect the position and size of the obstacles in the front. Second, we formulated the corresponding obstacle avoidance rules for different obstacles. Considering the continuity of obstacles, we divided the obstacles into five categories based on the location and size of obstacles in the field, as shown in Fig. 3. The five categories are as follows:

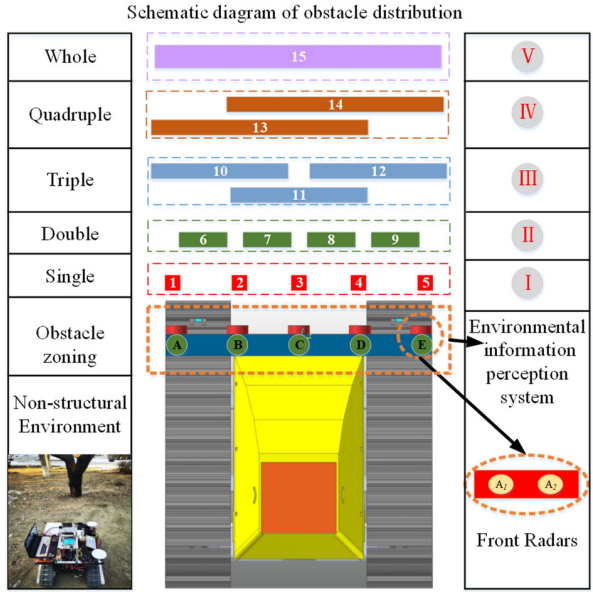

**Fig. 3.** Schematic diagram of the field obstacle classification based on obstacle information detection system.

Ⅰ: The vehicle adjusts its heading angle according to the position of the obstacle at this time when the obstacle at the scene is detected by only one group of the ultrasonic range radar, as shown in cases 1–5 in Fig. 3.

Ⅱ: The vehicle adjusts its heading angle according to the position of the obstacle at this time when the obstacle at the scene is detected only by two sets of the ultrasonic range radar, as shown in cases 6–9 in Fig. 3.

Ⅲ: The vehicle adjusts its heading angle according to the position of the obstacle at this time when the obstacle at the scene is detected by three sets of the ultrasonic range radar, as shown in cases 10–12 in Fig. 3.

Ⅳ: The vehicle adjusts its heading angle according to the position of the obstacle when an obstacle at the site is detected by four sets of the ultrasonic ranging radar, as shown in cases 13 and 14 in Fig. 3.

Ⅴ: The vehicle adjusts its heading angle according to the position of the obstacle at this time when the obstacle at the site is detected by five sets of ultrasonic range radar, as shown in case 15 in Fig. 3.

*2) Obstacle Avoidance Rule Design*

According to the requirement that the tracked intelligent transportation vehicle needs to maintain stable driving on non-structural roads, the vehicle dynamics constraints and the stability constraints in the obstacle avoidance process are comprehensively considered, and the vehicle lateral and longitudinal speed controls are combined for steering control. There are multiple avoidance paths when the vehicle encounters an obstacle. The vehicle not only needs to turn and avoid based on the direction of the obstacle, but also decelerate along the tangent direction of the obstacle. In the process of vehicle driving, according to engineering requirements, both sides of the vehicle should be reserved a certain safety distance $q$ ($q$ value should always be greater than 200 mm) to ensure that the vehicle can safely avoid obstacles. Fig. 4 shows the expert obstacle avoidance path process for tracked intelligent transport vehicles. When the tracked intelligent transport vehicle is driving on the road in the wild, the obstacle information detection system of the vehicle starts to work. On the other hand, the tracked intelligent transport vehicle maintains the original heading and driving to the target point when the vehicle-mounted radar does not detect the obstacle. When the vehicle-mounted radar detects the obstacle ahead, the vehicle adjusts the heading to avoid the obstacle. The size of the steering angle $\theta_m$ is related to the position of the obstacle.

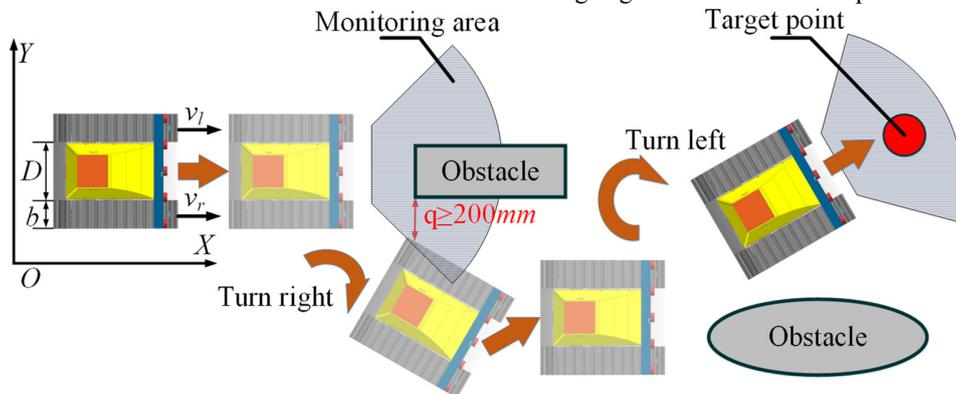

**Fig. 4.** Schematic diagram of the obstacle avoidance path of the tracked intelligent transport vehicle.

The forward and steering angular velocities of the unmanned vehicle at moment $t$ can be expressed as [32]-[34].

$$\begin{cases} \dot{x}(t) = \frac{1}{2}(v_l + v_r)\cos\theta_d \\ \dot{y}(t) = \frac{1}{2}(v_l + v_r)\sin\theta_d \\ \dot{\theta}_d(t) = \frac{-v_l + v_r}{D + 2b} \end{cases} \quad (1)$$

Then, the equation of the state of motion of the unmanned vehicle can be expressed as follows:

$$\dot{W} = \begin{bmatrix} \dot{x} \\ \dot{y} \\ \dot{\theta}_d \end{bmatrix} = \begin{bmatrix} \cos\theta_d & 0 \\ \sin\theta_d & 0 \\ 0 & 1 \end{bmatrix} \begin{bmatrix} v_o \\ \omega_o \end{bmatrix}$$

$$= H \begin{bmatrix} v_o \\ \omega_o \end{bmatrix} \quad (2)$$

$$= \begin{bmatrix} \frac{1}{2}\cos\theta_d & \frac{1}{2}\cos\theta_d \\ \frac{1}{2}\sin\theta_d & \frac{1}{2}\sin\theta_d \\ -\frac{1}{D+2b} & \frac{1}{D+2b} \end{bmatrix} \begin{bmatrix} v_l \\ v_r \end{bmatrix}$$

The turning radius $R$ of the track is given by

$$R = \frac{(2v_r + \Delta v)(D + 2b)}{2\Delta v} \quad (3)$$

The heading angle of the unmanned vehicle can be calculated using Eq. (4), where $i=1, 2, \ldots 5$ and $m=1, 2, \ldots 15$.

$$\theta_m = \arctan\frac{q + R_i}{1500} \quad (4)$$

After calculating the above formula, reserving a certain safety margin for the steering angle of the tracked intelligent transport vehicle, and considering various engineering factors, the actual steering angle of the unmanned vehicle can be determined. Table I lists the calculation results of $\theta_m$.

TABLE I
OBSTACLE AVOIDANCE STRATEGIES FOR OBSTACLES IN DIFFERENT LOCATIONS

| Condition | A | B | C | D | E | $\theta_m$ (°) | $\omega_L$ (rad/s) | $\omega_R$ (rad/s) | $\omega_O$ (rad/s) |
|---|---|---|---|---|---|---|---|---|---|
| 1 | 1 | 0 | 0 | 0 | 0 | -6 | 1.5 | 0.5 | 1 |
| 2 | 0 | 1 | 0 | 0 | 0 | -20 | 4.5 | 1.6 | 3.05 |
| 3 | 0 | 0 | 1 | 0 | 0 | -28 | 6.8 | 2.4 | 4.6 |
| 4 | 0 | 0 | 0 | 1 | 0 | +20 | 1.6 | 4.5 | 3.05 |
| 5 | 0 | 0 | 0 | 0 | 1 | +6 | 0.5 | 1.5 | 1 |
| 6 | 1 | 1 | 0 | 0 | 0 | -20 | 4.5 | 1.6 | 3.05 |
| 7 | 0 | 1 | 1 | 0 | 0 | -28 | 6.8 | 2.4 | 4.6 |
| 8 | 0 | 0 | 1 | 1 | 0 | +28 | 2.4 | 6.8 | 4.6 |
| 9 | 0 | 0 | 0 | 1 | 1 | +20 | 1.6 | 4.5 | 3.05 |
| 10 | 1 | 1 | 1 | 0 | 0 | -28 | 6.8 | 2.4 | 4.6 |
| 11 | 0 | 1 | 1 | 1 | 0 | -45 | 10.8 | 3.9 | 7.35 |
| 12 | 0 | 0 | 1 | 1 | 1 | +28 | 2.4 | 6.8 | 4.6 |
| 13 | 1 | 1 | 1 | 1 | 0 | -42 | 10 | 3.6 | 6.8 |
| 14 | 0 | 1 | 1 | 1 | 1 | +42 | 3.6 | 10 | 6.8 |
| 15 | 1 | 1 | 1 | 1 | 1 | -65 | 15.6 | 5.6 | 10.6 |

$\theta_m$ is the actual steering angle of the tracked intelligent transportation vehicle for obstacle avoidance, "-" means the tracked intelligent transportation vehicle turns right, "+" means the tracked intelligent transportation vehicle turns left, $\omega_s$ ($\omega_s$=5.5 rad/s) is the initial drive wheel speed of both sides of the crawler intelligent transportation vehicle during the straight running condition, $v_s$ ($v_s$=0.9 m/s) is the initial linear velocity of the vehicle, $\omega_L$ and $\omega_R$ are the rotational speeds of the left and right track driving wheels, respectively, and $\omega_O$ linear velocity of the geometric center point of the vehicle track. To sum up, the flow of vehicle autonomous obstacle avoidance algorithm based on non-structural road is shown in algorithm I.

**Algorithm I:** Expert driving obstacle avoidance strategy algorithm

    Initialize environment settings (vehicle and obstacles);
    Set safe obstacle avoidance distance for vehicle $L_T$=1500*mm*;
    **Data Set:** $I_s$ (As shown in III·A·1))
    **Input:** Obstacle information and vehicle status parameters
    **Output**: Heading angle and vehicle speed

1. Determine the three-dimensional coordinates of the target point (*x*, *y*, *z*);
2. Tracked intelligent transport vehicle starts driving;
3. Record the detection data of the ultrasonic sensor;
4. Multi-sensor information fusion;
5. **for each** $i \in [A, B, ..., E]$ **do**
6.    $L_i \leftarrow$ Ranging value of each group of ultrasonic radar sensors (after fusion);
7.    **if** $L_i \leq L_T$ **then**
8.       step1: Classify obstacles (Figure 3);
9.       step2: Adjust the speed of the tracks on both sides ($\omega_L$, $\omega_R$) according to the obstacle avoidance strategy in Table I;
10.      step3: Drive at speed $v_s$ to avoid obstacles;
11.      step4: Vehicle returns to the initial course;
12.    **else**
13.       Keep the original course and drive to the target point;
14.    **end**
15.    Keep driving;
16.    **if** $L_i \leq L_T$ **then**
17.       Continue to implement the above obstacle avoidance strategy;
18.    **else**
19.       Continue driving to the target point;
20.    **end**
21. **end**
22. The vehicle continues to the target point;
23. Using the system to collect operation data and experience.

**Algorithm. I.** The algorithm flow of the vehicle autonomous obstacle avoidance control based on non-structural road.

*B. Vehicle Stability Monitoring*

This study needs to analyze and evaluate the dynamic stability of the vehicle to improve the driving stability of the vehicle. Dynamic stability is an important criterion for evaluating vehicle stability. A vehicle is judged whether it is in a stable state through the analysis of stability indicators, and then early warning and active anti-roll control strategy are implemented. There are various reasons for the instability of the vehicle, such as climbing and rolling over, road collapse, side slip and tripping, and fast turning when encountering obstacles. Regardless the cause of vehicle instability, there must be some common features before this occurs, such as the lateral acceleration of the vehicle, transverse angular acceleration, lateral angular velocity, lateral tilt angle, and other parameters that have exceeded a certain threshold value. Therefore, in this study, we employed IMU to collect vehicle state parameters, such as three-axis attitude angle, three-axis angular velocity, and centroid dynamic position during driving.

A certain safety distance margin needs to be maintained between the vehicle and the edge of the obstacle, while other suboptimal paths have long distances and too many sharp turns, resulting in a large yaw rate acceleration and poor stability. Therefore, our proposed stability control rule aims to find the best path to avoid the obstacles ahead while maintaining body stability. Based on the slope information and the driving information of the vehicle itself, the dynamic stability index of tracked intelligent transportation vehicle can be expressed as [35] [36]:

$$DSE_{\min} = \frac{\beta}{\beta_{\max}} \begin{pmatrix} \lambda_1 a_x + \lambda_2 a_y + \lambda_3 a_z + \\ \lambda_4 a_{rx} + \lambda_5 a_{ry} + \lambda_6 a_{rz} \end{pmatrix} \quad (5)$$

In Eq. (5), $β_{max}$ is the maximum road climbing angle of the tracked intelligent transport vehicle, $a_x$, $a_y$, and $a_z$ are the lateral acceleration, forward acceleration, and longitudinal acceleration of the tracked intelligent transport vehicle, respectively, $a_{rz}$, $a_{ry}$, and $a_{rx}$ are the pitch angle acceleration, roll angle acceleration, and heading angle acceleration of the tracked intelligent transport vehicle, respectively. This evaluation index integrates the vehicle motion parameters and road slope angle. Moreover, $λ_1$, $λ_2$, $λ_3$, $λ_4$, $λ_5$ and $λ_6$ are the weight coefficients.

$$λ_1 + λ_2 + λ_3 + λ_4 + λ_5 + λ_6 = 1 \quad (6)$$

Through our experiments, we have obtained the forward acceleration, lateral acceleration, heading angular acceleration, and roll angular acceleration of the vehicle that will have a great impact on the stability of the vehicle. Therefore, the weight coefficient is set to: $λ_1=λ_2=λ_5=λ_6=0.2$ and $λ_3=λ_4=0.1$. Table II lists the stability index constraints.

TABLE II
INSTABILITY EVALUATION INDEX TABLE OF TRACKED INTELLIGENT TRANSPORTATION VEHICLE

| Evaluating indicator | $a_x$ | $a_y$ | $a_z$ | $a_{rx}$ | $a_{ry}$ | $a_{rz}$ | $β$ | DSE |
|---|---|---|---|---|---|---|---|---|
| Value | ≤1g | ≤0.8g | ≤2g | ≤9 rad/s² | ≤13 rad/s² | ≤16 rad/s² | ≤20° | 7.26 |

This study used Attention-long short-term memory (Attention-LSTM) based on deep learning method to adjust the driving control strategy of the vehicle based on the processed obstacle information and vehicle driving data and ensure the driving stability during obstacle avoidance.

*C. Deep Learning Model Based on Attention-LSTM*

It is necessary to use the obstacle information detection system to process and analyze the obstacle information in real time due to the relatively complex environment of non-structural roads. Considering that the vehicle central control unit has no reference track and off-line map, it is difficult to make various expert experience judgments on obstacles.

Deep learning has powerful complex system characterization capability, big data processing capability and automatic feature extraction capability, which is feasible and superior in detection tasks. Based on these, this study proposed a multi-level obstacle avoidance prediction network model, which is based on LSTM (long short-term memory) model and encoder-decoder structure [37]-[39]. A multi-level obstacle avoidance mechanism was introduced for the first time to model the dynamic spatiotemporal correlation between the sensors of the unmanned tracked vehicle to address the problem of high stability obstacle avoidance.

Based on the RNN model, LSTM adds input gate, forgetting gate, output gate, and cell unit, as shown in Fig. 5. Among them, three control gates are responsible for controlling the inflow and outflow of information, thereby protecting and controlling the state of the cell unit, which is responsible for remembering the information of the previous moment. From Fig. 5, the forward propagation formula of LSTM unit can be deduced as follows:

$$z_t^j = \sum_{i=1}^{I} w_{ij} x_i^t + \sum_{h=1}^{H} w_{hj} s_h^{t-1} + \sum_{c=1}^{C} w_{cj} m_c^{t-1} + b_j \quad (7)$$

$$s_j^t = f(z_j^t) \quad (8)$$

$$z_k^t = \sum_{i=1}^{I} w_{ik} x_i^t + \sum_{h=1}^{H} w_{hk} s_h^{t-1} + \sum_{c=1}^{C} w_{ck} m_c^{t-1} + b_k \quad (9)$$

$$s_k^t = f(z_k^t) \quad (10)$$

$$z_c^t = \sum_{i=1}^{I} w_{ic} x_i^t + \sum_{h=1}^{H} w_{hc} s_h^{t-1} + b_c \quad (11)$$

$$m_c^t = s_k^t m_c^{t-1} + s_j^t g(z_c^t) \quad (12)$$

$$z_l^t = \sum_{i=1}^{I} w_{il} x_i^t + \sum_{h=1}^{H} w_{hl} s_h^{t-1} + \sum_{c=1}^{G} w_{cl} m_c^t + b_l \quad (13)$$

$$s_h^t = s_l^t \phi(m_c^t) \quad (14)$$

LSTM can save historical information, which inherits the advantages of the traditional neural network and excavates historical time data [40] [41]. The traditional LSTM converts the input sequence into a fixed length vector and saves all the information, limiting the model memory, and is easy to lose information when dealing with long sequence problems. A single LSTM model converts the input sequence into a fixed-length vector and saves all the information. Moreover, it cannot detect the important parts that affect the current obstacle avoidance strategy, reducing the utilization of information. In this study, the excessive rolling angular acceleration and lateral acceleration of the vehicle can cause vehicle instability according to the experience. The addition of attention mechanism can make up for this defect. It can give weight to different information and strengthen the memory of important information. This study attempts to combine LSTM with the attention mechanism for high stability obstacle avoidance strategy prediction. The overall structure of the Attention-LSTM prediction model proposed in this study is shown in Fig. 5.

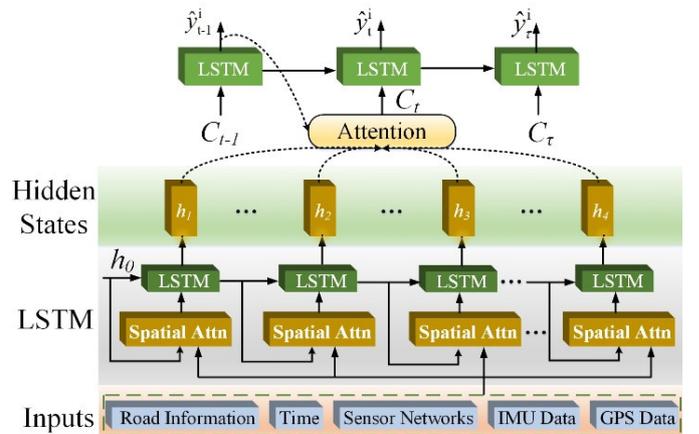

**Fig. 5.** Structure diagram of multi-level obstacle avoidance network model based on attention mechanism.

*D. Design of Stability Obstacle Avoidance Control Framework*

This study proposed a stabilization and obstacle avoidance control strategy based on deep learning modeling using Attention-LSTM deep learning method to realize the dynamic prediction of vehicle stabilization and obstacle avoidance strategy under the harsh environment in the field. Fig. 6 shows the schematic diagram of the control strategy, and the framework consists of three parts.

Part I: Vehicle obstacle avoidance strategy. We developed the corresponding expert bypass strategy based on different obstacle information to find the optimal safe path to avoid the obstacles ahead under the condition of the shortest path. First, an obstacle classification strategy based on obstacle characteristics and spatial relative position relationship was proposed, and the obstacle information perception system was used to obtain the obstacle information in front. Second, an optimal obstacle avoidance strategy was designed to avoid obstacles on the vehicle route. Unmanned vehicles avoid collision between vehicles and obstacles by automatically adjusting the driving attitude of vehicles. The vehicle obstacle avoidance strategy is shown in part I of Fig. 6.

Part II: Vehicle stability judgment. First, the vehicle used IMU to collect the motion state information during driving (including three-axis attitude angle, three-axis angular velocity, and centroid position), in which the vehicle pitch angle collected by IMU can be used to calculate the slope angle of the current slope. Subsequently, the dynamic stability of the vehicle was summarized and analyzed by comparing the changes of various parameters during vehicle driving. The vehicle stability judgment is shown in part II of Fig. 6.

Part III: Learn expert obstacle avoidance strategies. To find the optimal safe path to avoid the obstacles ahead under the condition of the shortest path, this study developed the corresponding expert bypassing strategy based on different obstacle information and used Attention-LSTM to adjust the optimal obstacle avoidance strategy of the vehicle based on the processed obstacle information and vehicle driving data. Attention-LSTM is a new recurrent neural network structure based on gradient learning algorithm, which solves the problem of gradient disappearance and gradient explosion during the training of long sequences. This structure has certain advantages in sequence modeling and has long-term memory function, which will be applied in this study. This strategy has the features of simple logic and easy implementation. In this study, we used this feature to predict the obstacle avoidance strategy of tracked intelligent transportation vehicles. Learn expert obstacle avoidance strategies are shown in part III in Fig. 6.

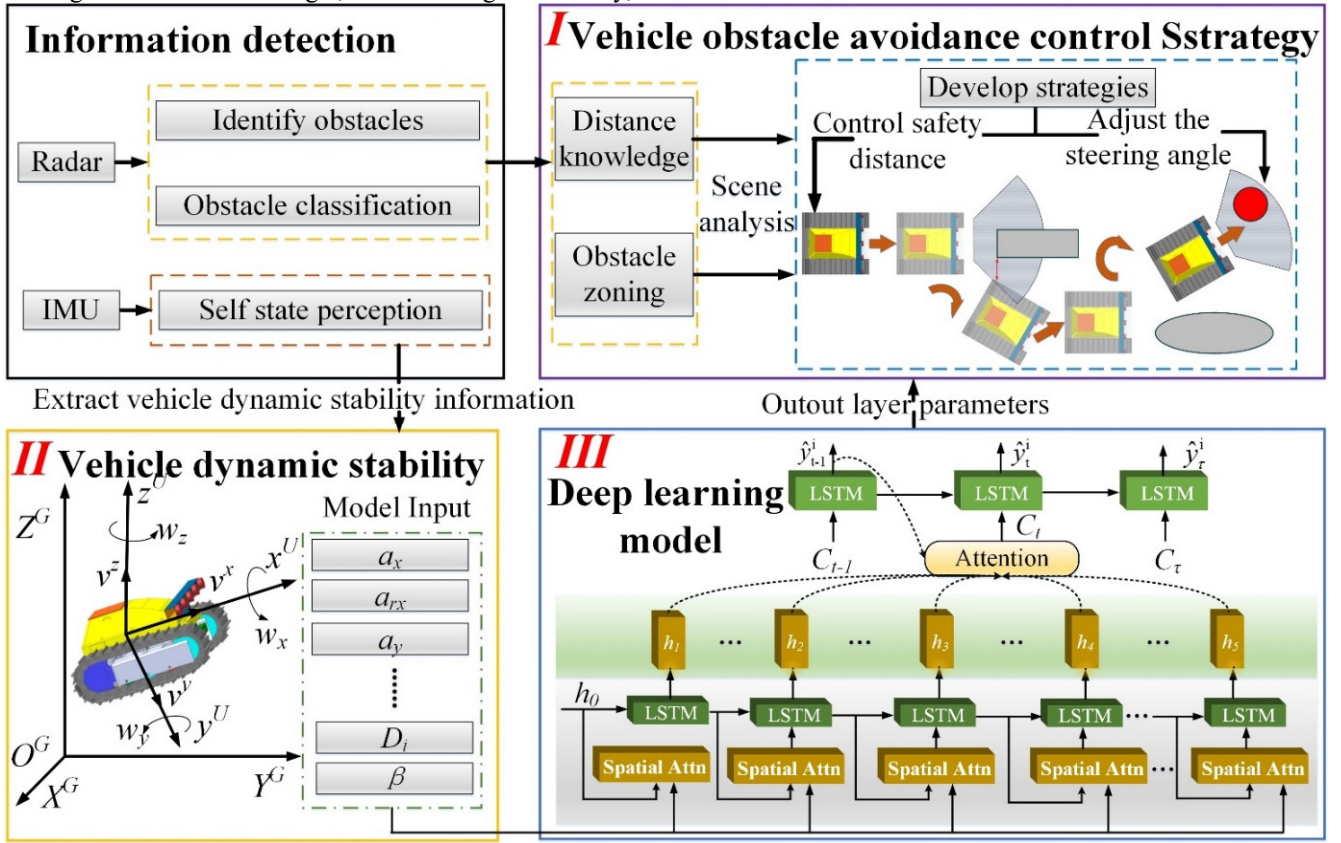

**Fig. 6.** Design of stability obstacle avoidance control strategy framework.

IV. SIMULATION AND EXPERIMENT

Deep learning has been successfully applied to many engineering fields, and a common feature of these applications is the need to use large amounts of feature data to train deep learning models. The working environment of tracked intelligent transport vehicles is relatively harsh, and the instability conditions of tracked intelligent transport vehicles are relatively complex in field experiments. However, it is not an economical and practical method to wear the tracked intelligent transportation vehicle only to verify the obstacle avoidance control strategy. Therefore, we used virtual prototype simulation to establish the data set. The effectiveness of this method needs to be fully verified in simulation experiments before large-scale field experiments can be carried out.

*A. Simulation Settings*

The speed and heading angle of the tracked intelligent transport vehicle are planned in real time by the central computing unit, as shown in Fig. 7. We used a virtual prototype model to acquire data sets based on the relevant design parameters of the tracked intelligent transportation vehicle since dangerous situations, such as vehicle rollover, may occur during the experiment. Then, we conducted ground experiments to verify the effectiveness and stability of the method.

The relevant technical parameters of the tracked intelligent transport vehicle designed in this study are shown in Table III.

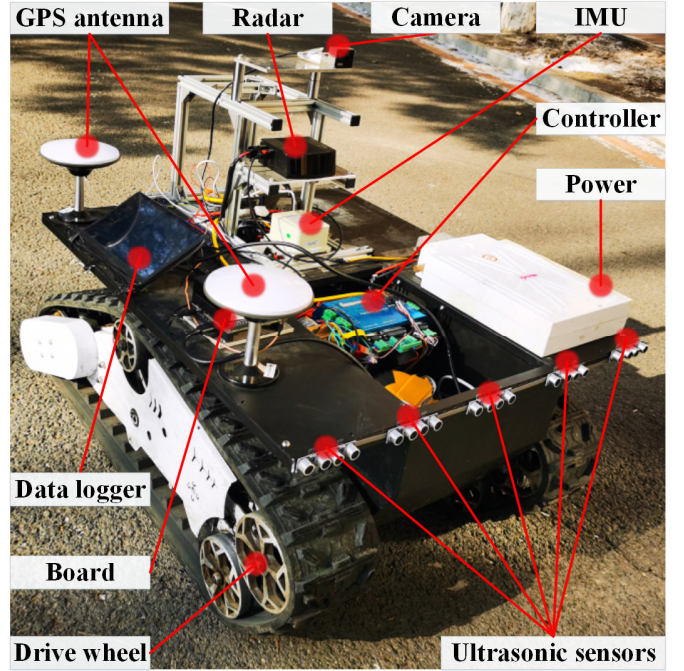

**Fig. 7.** Structural diagram of tracked intelligent transport vehicle.

TABLE III
DESIGN PARAMETERS OF UNMANNED VEHICLE CHASSIS FOR TEST

| Symbol | Parameter | Value | Units |
| --- | --- | --- | --- |
| $m$ | Chassis quality | 450 | Kg |
| $b$ | Gauge | 1 | m |
| $D$ | Body width | 1.25 | m |
| $L_l$ | Total length of tracked chassis | 1.2 | m |
| $l$ | Track ground length | 0.881 | m |
| $H$ | Track height | 0.331 | m |
| $V_{max}$ | Maximum travel speed | 2.8 | m/s |
| $i$ | Reduction ratio | 100 | m |
| $T$ | Main traction force | 500 | Kg |
| $\beta$ | Maximum climbing angle | 30 | Angle |
| $P$ | Ground pressure | 0.045 | Mpa |

*B. Simulation Experiment*

In this section, we used the vehicle dynamics software Recurdyn and the simulation software Matlab/Simulink for joint simulation, establishment of the driving control model of tracked intelligent transportation vehicle in Simulink, and development of the kinematics and dynamics model of tracked intelligent transportation vehicle in Recurdyn According to the above design requirements, the Matlab/Simulink-Recurdyn joint simulation model of the active control system is shown in Fig. 8.

We referred to relevant literature and set the following ground contact parameters, as shown in Table IV, to better analyze the driving process of unmanned vehicles.

TABLE IV
GROUND CONTACT PARAMETER SETTING

| Parameter | Values |
| --- | --- |
| Terrain Stiffness | 4.7613e-004 |
| Cohesion | 1.04e-003 |
| Shearing Resistance Angle | 28 |
| Shearing Deformation Modulus | 25 |
| Sinkage Ratio | 5.e-002 |

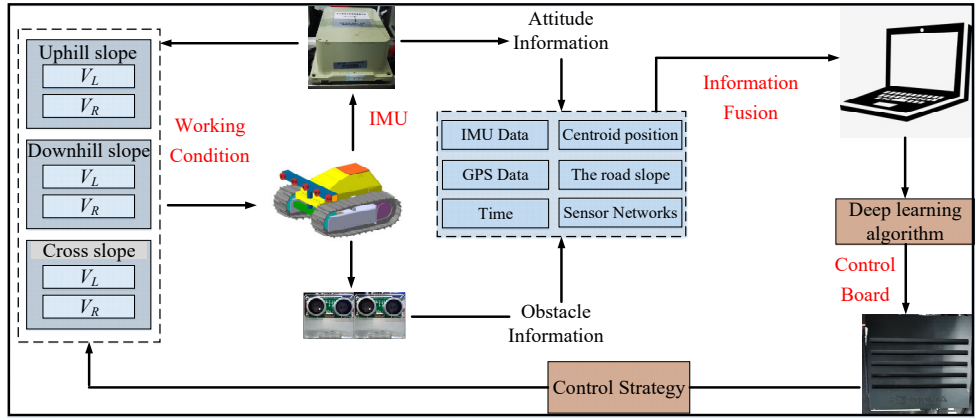

**Fig. 8.** Composition diagram of active control system.

## C. Establishment of Data Set

A virtual prototype was used for joint simulation to obtain the training data set. We preset 15 kinds of obstacles in the simulation environment and conducted simulation experiments according to different slope conditions and obstacle distributions. In this section, a total of 500 simulation experiments were performed and 5000 sets of valid data were selected for training (each set contains vehicle motion state parameters and corresponding sensor data). Fig. 9 shows some obstacle avoidance conditions and the vehicle three-axis attitude angle data. Among them, (a) is the obstacle avoidance condition of the unmanned vehicle on the flat road, (b) is the obstacle avoidance condition of the unmanned vehicle climbing, (c) is the obstacle avoidance condition of the unmanned vehicle downhill, and (d) is the obstacle avoidance condition of the unmanned vehicle longitudinal slope. From Fig. 9, the sensor data changes relatively smoothly, and the current slope information can be obtained using the roll and pitch angles of the unmanned vehicle. The collected vehicle data include obstacle distribution, climbing angle, three-axis attitude angle, dynamic position of mass center, three-axis acceleration, and speed of the crawler driving wheels on both sides. The location information of the obstacles and the state information of the vehicle are set as the inputs to the prediction model to provide training data.

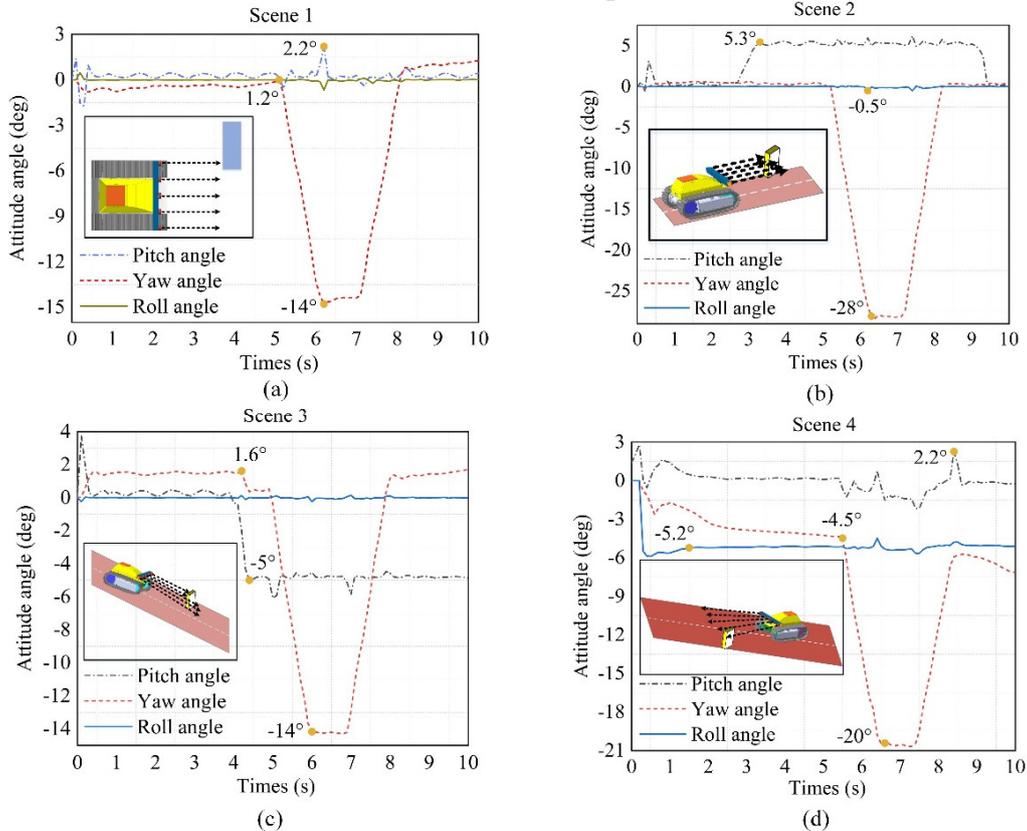

**Fig. 9.** Schematic diagram of the three-axis attitude angle in four typical obstacle avoidance scenarios during training (a)(b)(c)(d). (a) Obstacle avoidance condition of the unmanned vehicle on the flat road. (b) Obstacle avoidance condition of the unmanned vehicle uphill. (c) Obstacle avoidance condition of the unmanned vehicle downhill. (d) Obstacle avoidance condition of the unmanned vehicle on the longitudinal slope.

## V. RESULTS AND DISCUSSION

### A. Results of Attention-LSTM and Comparison with Other Models

According to the input of different ground slope information, we divided obstacle avoidance conditions into two types, such as based on uphill and downhill obstacle avoidance and longitudinal slope obstacle avoidance, and two deep learning networks were trained. In this study, we set the initial learning rate to 0.005, batch size to 20, and Maxepochs to 2000. RMSE and loss during training are shown in Fig. 10. For ease of understanding, the coordinates of RMSE (right ordinate) increase from top to bottom since both RMSE and loss decrease with the increase of iteration times. The RMSE has reached a relatively low level in a short time. The results show that there is no overfitting or underfitting phenomenon in the network, which shows that the model has good generalization ability.

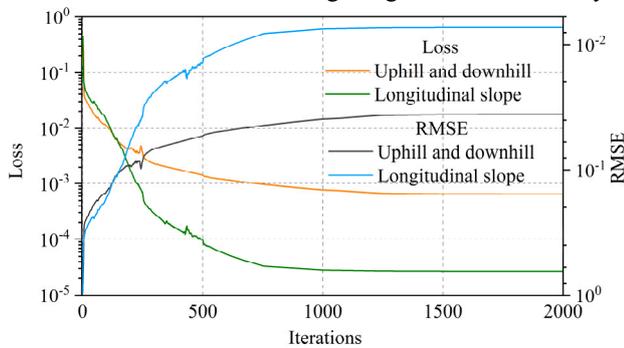

**Fig. 10.** RMSE and Loss of the Attention-LSTM.

Based on Fig. 11, the RMSE and the average relative error rate of the model gradually decrease as the slope decreases, indicating that the regression performance of the model improves as the slope decreases. The maximum average relative error rate reaches 19.3% when the slope is 15°, which may be due to the rapid change of driving information and complex vehicle conditions under large angle maneuvering conditions. The average relative error rate of the deep learning model established in this study is 15%, which meets the current engineering requirements.

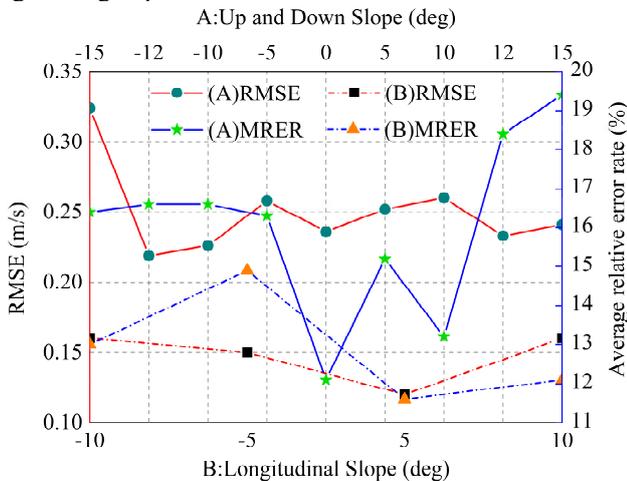

**Fig. 11.** Root mean square error and average relative error rate diagrams of the proposed model.

### B. Comparison with Other Models

To compare the prediction accuracy of the deep learning algorithms, we used five other popular networks to perform a comprehensive performance evaluation with the model properties proposed in this study and evaluated the performance of the models using the average relative error rate and loss, as shown in Fig. 12 and Table V.

Based on the violin plot (Fig. 12), the accuracy of our proposed model is far better than other networks, especially the BP neural network, whose relative average relative error rate has reached 34.11% with relatively low accuracy. The average relative error rate of LSTM, RNN, and Gru neural networks has exceeded 20%. Although the average relative error rate of LSTM-CNN neural network is 18.97%, its error value is more dispersed than the model used in this study. Based on the comparison results, the model used in this study has better accuracy and reliability.

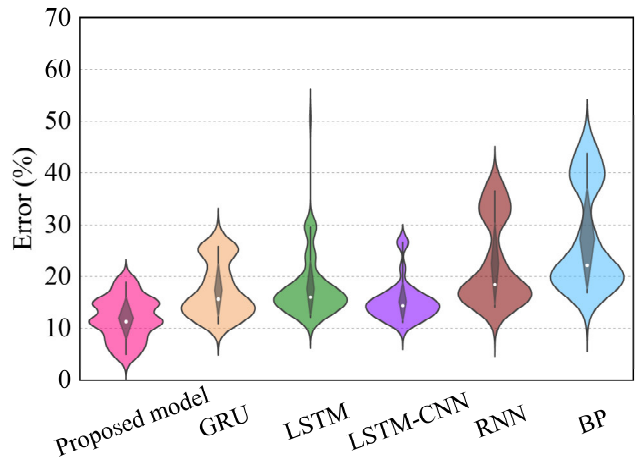

**Fig. 12.** Violin figure of average relative error for each model.

In terms of loss, our model shows a lower loss value compared than that of other models, as shown in Table V. Clearly, our model is superior to other models in terms of loss.

TABLE V
PERFORMANCE OF EACH FEATURE EXTRACTION MODEL

| Base Model | Average relative error (%) | Loss |
|---|---|---|
| GRU | 21.77 | 0.053 |
| LSTM | 22.09 | 0.091 |
| LSTM-CNN | 18.97 | 0.032 |
| RNN | 28.12 | 0.179 |
| BP | 34.11 | 0.237 |
| **Proposed model** | **14.95** | **0.027** |

### C. Preliminary Comparison Between Virtual Simulation and Ground Experiment

We conducted simulation and preliminary ground experiments to validate the flat road obstacle avoidance conditions and verify the accuracy of the proposed method. Under the conditions of the same obstacle locations and sizes in the virtual and real environments, we used the method proposed in this study to predict the control results of the tracked intelligent transport vehicle separately and compared the simulation results with the real ground experiment results.

Fig. 13 shows the comparison of the control results after starting the unmanned vehicle under the flat road obstacle avoidance condition. Based on Fig. 13(a), in the real environment, the speed gradually reaches a stable state after the unmanned vehicle starts. The starting speed is different from that in the real environment since the unmanned vehicle starts at a constant acceleration in the virtual environment. We selected the speed data of the unmanned vehicle when it reaches the steady state and calculated the error. The average relative error rate of the centroid speed of the unmanned vehicle in the virtual environment and the real environment is 6.5%, which meets the requirements of this study.

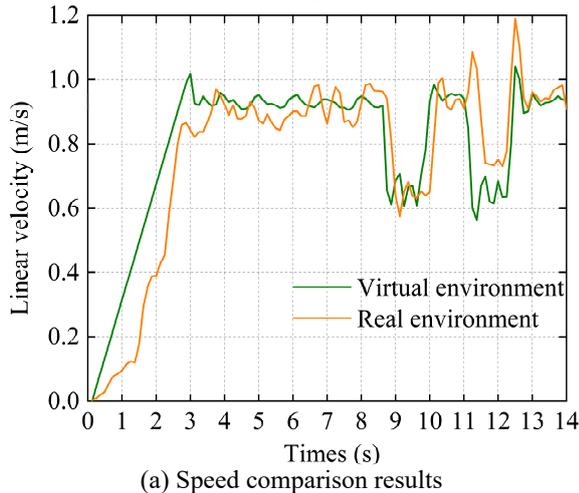

(a) Speed comparison results

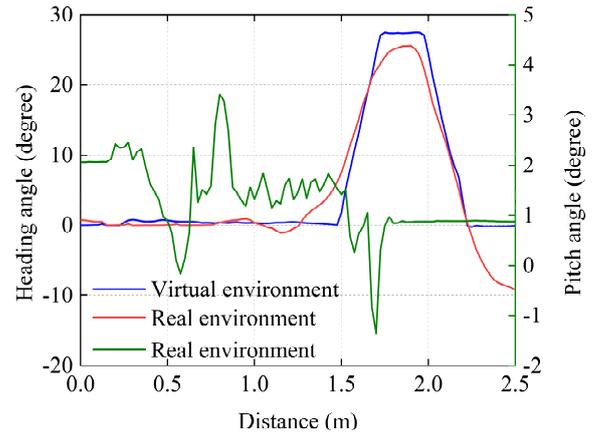

(b) Heading angle comparison results

**Fig. 13.** Preliminary comparison between virtual simulation and real ground experiment (under obstacle avoidance condition on flat road).

Fig. 13(b) shows the comparison of the heading angle of the unmanned vehicle after startup. Based on the figure, the maximum heading angle error of the unmanned vehicle in virtual and real environments is 2°. Considering the relative turbulence of the ground environment (see the actual pitch angle in Fig. 13(b)), control errors, and other factors, the heading angle error meets the engineering requirements.

Through the comparison and analysis of the experimental results of the virtual and real environments, it is proven that the method proposed in this study has good prediction effect and high stability and can be used for the next slope obstacle avoidance experiment.

*D. Real World Tests*

To verify the accuracy of the method proposed in this study, we conducted field experiments with large slopes for the proposed slope obstacle avoidance conditions. The following is our strategy for deploying work in the actual field environment.

First, we carried out the field ground layout. Fig. 14 shows the test ramp. The length of the test section and the ramp section must not be less than 8 and 5 m, respectively. There are transition sections in front and behind the test section, and the flat and straight section in front of the slope must not be less than 3 m. The road is an asphalt pavement with dry, solid surface, and uniform slope. Then, we arranged objects such as foam boxes and masonry as obstacles on the non-structural gradient pavement, and natural obstacles such as wild stones were also present on the pavement. According to the engineering requirements, the tracked intelligent transport vehicle stops on the straight road section close to the ramp area, after starting and gradually driving into the climbing road, it needs to avoid the obstacles ahead and complete the body back at an appropriate speed.

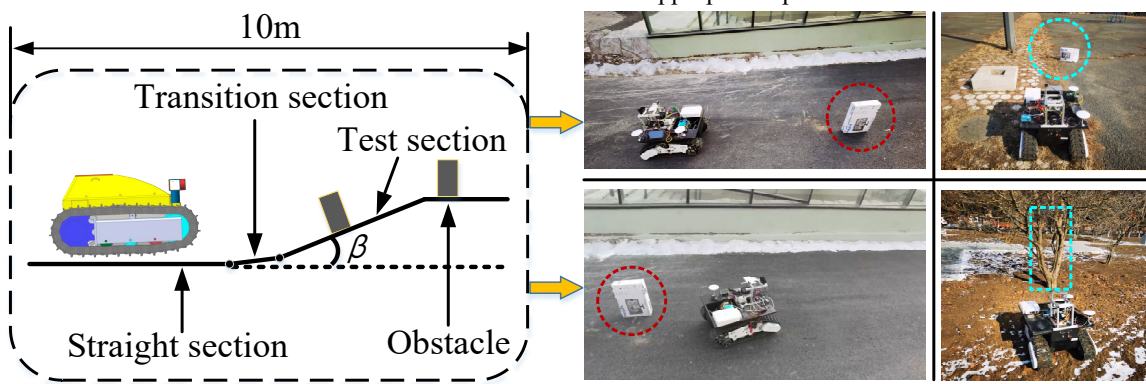

**Fig. 14.** Obstacle avoidance experiment of driverless vehicle in field environment (the left figure is the environmental diagram, and the right figure is the real environment diagram).

We evaluated the success rate of obstacle avoidance by recording the driving trajectory and IMU data of unmanned vehicles in the test section (obstacle avoidance completed on 71/72) to demonstrate the effectiveness and feasibility of the method in a real field environment. The experiment is successful if the unmanned vehicle successfully avoid obstacles and does not lose stability during driving. Note that in this experiment, IMU adopted the northeast sky coordinate system (i.e., $y$-axis-pointing north; $z$-axis-pointing skyward; $x$-axis-pointing eastward). For easier understanding, we transformed the coordinate system accordingly.

Scenario 1: Scenario 1 is the climbing condition of unmanned vehicles, such as in case 4 (see Table I).

The variation of the three-axis motion attitude angle of the unmanned vehicle is shown in Fig. 15(a). Based on this figure, the unmanned vehicle performs obstacle avoidance in the process of climbing a 9° slope and it decreases its speed to the left to avoid the obstacle by 21.7°. The driving path and obstacles of the unmanned vehicle are shown in Fig. 15(b), in which the unmanned vehicle successfully plans a safe obstacle avoidance path and steers to avoid obstacles at the 4 s. The three-axis attitude angular acceleration and three-axis acceleration of the unmanned vehicle during driving are shown in Fig. 15(c). From Table VI, the maximum yaw angle acceleration of the unmanned vehicle during obstacle avoidance is -5 rad/s$^2$ and the maximum lateral acceleration is 0.5 g. The results show that the stability index of unmanned vehicle in obstacle avoidance process is less than the critical value of instability. The obstacle avoidance process is more efficient and stable based on our proposed method.

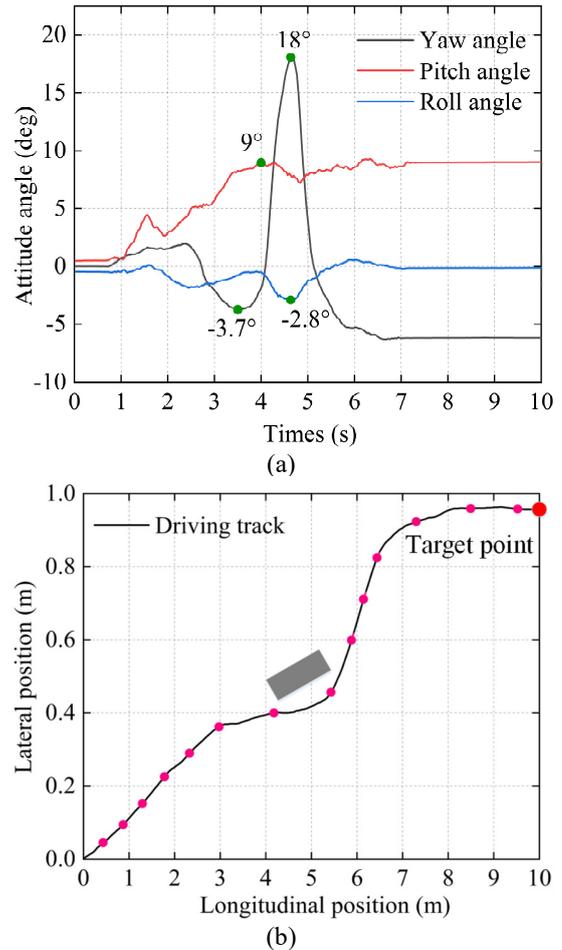

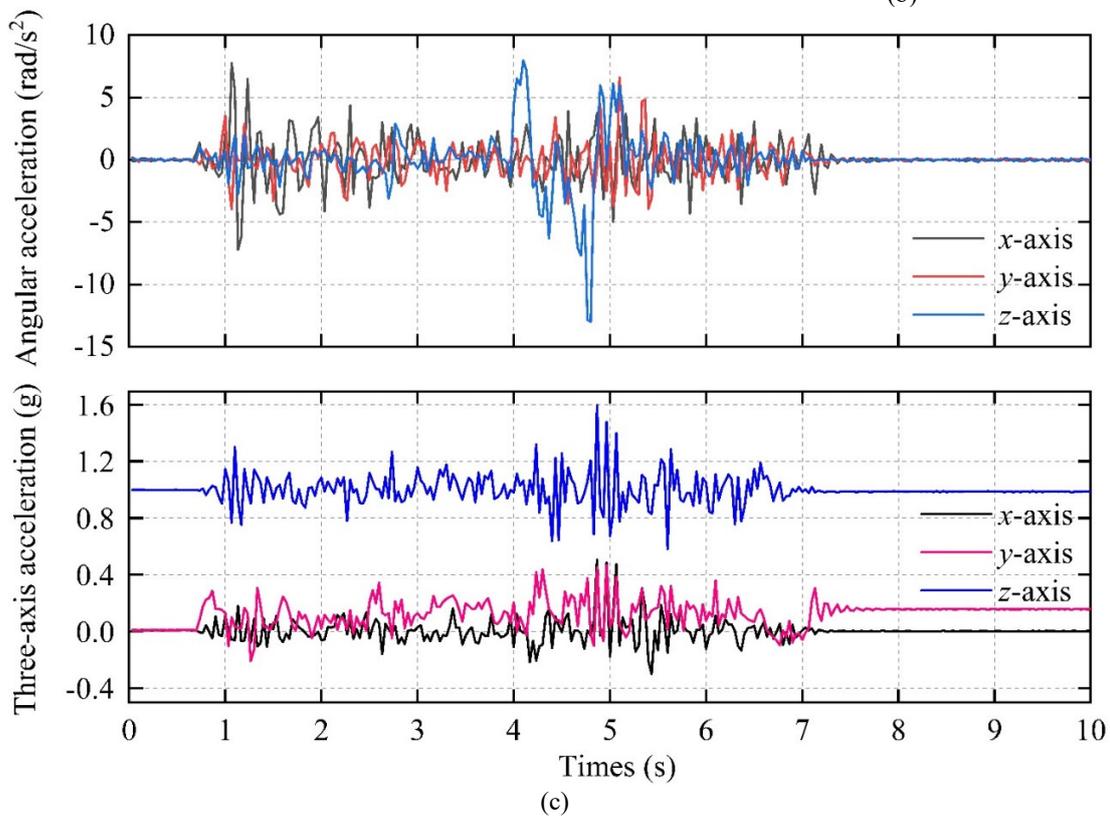

**Fig. 15.** Experimental result diagram of Scenario 1 (climbing condition).

Scenario 2: Scenario 2 is when the unmanned vehicle performs a downhill working condition, similar with case 14 (see Table I).

The variation of the three-axis motion attitude angle of the unmanned vehicle is shown in Fig. 16(a). Based on this figure, the unmanned vehicle is descending and performing obstacle avoidance in a 7° slope. In addition, it reduced its speed to 43° to the left to avoid the obstacle. The travel path and obstacles of the unmanned vehicle are shown in Fig. 16(b). From this figure, the unmanned vehicle successfully plans a safe obstacle avoidance path and makes a steering avoidance at 4.5 s. The three-axis attitude angular acceleration and three-axis acceleration of the unmanned vehicle during travel are shown in Fig. 16(c). According to Table VI, the maximum yaw angular acceleration is -12 rad/s$^2$ when the unmanned vehicle avoids obstacles, and the maximum lateral acceleration is 0.47 g. The results show that the stability index in the obstacle avoidance process is less than the instability critical value, and the obstacle avoidance process of the unmanned vehicle is more efficient and stable based on our proposed method.

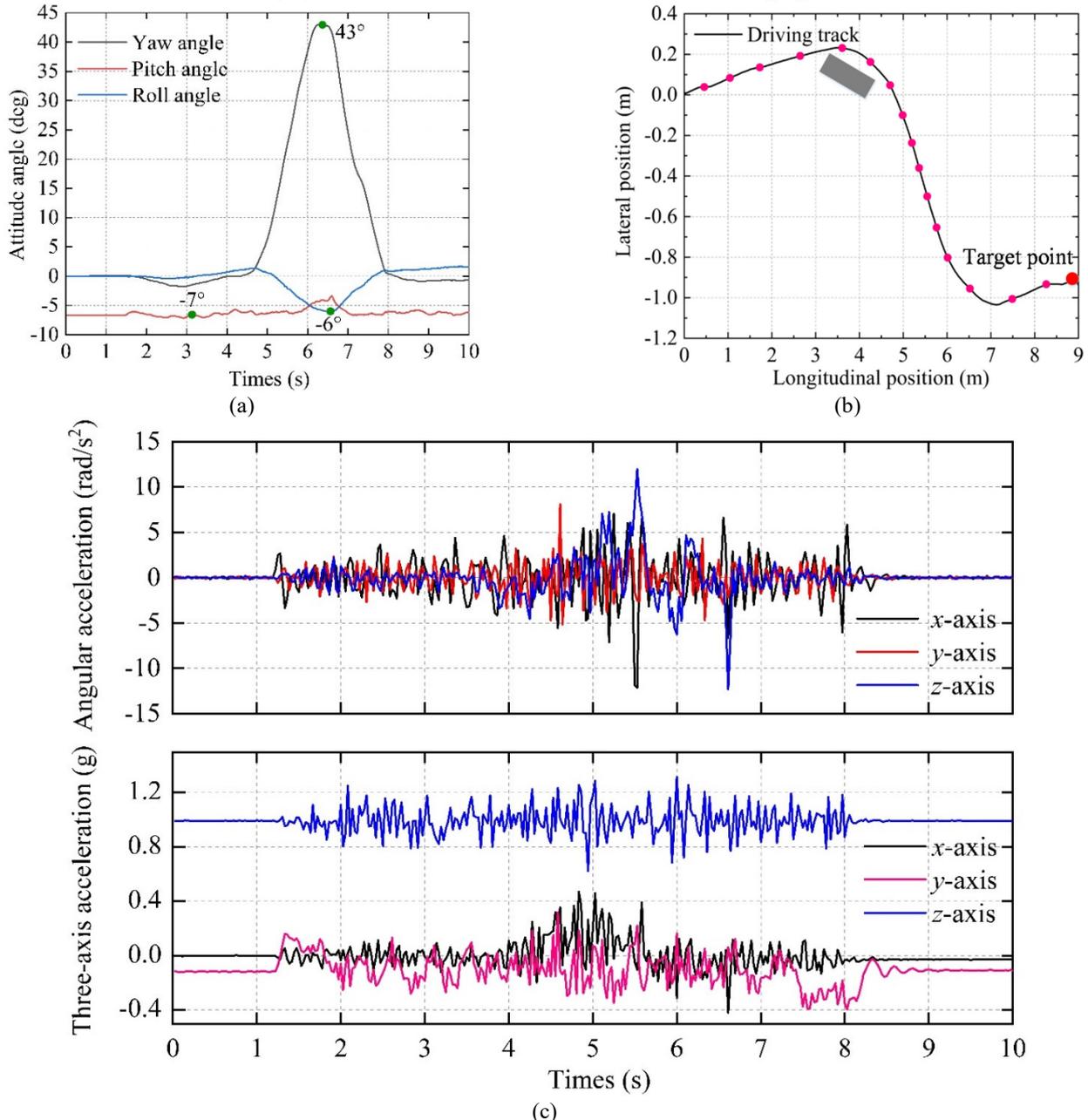

**Fig. 16.** Experimental result diagram of Scenario 2 (downhill condition).

Scenario 3: Scenario 3 is when the unmanned vehicle performs a longitudinal slope condition, similar in case 15 (see Table I).

The variation of the three-axis motion attitude angle of the unmanned vehicle is shown in Fig. 17(a). Based on this figure, the unmanned vehicle performed the obstacle avoidance condition on the longitudinal slope (the slope is -6°) at this time, and the unmanned vehicle reduced the speed and turned to the left at 71° to avoid the obstacle (for ease of understanding, the right side ordinate is the heading angle). The driving path and obstacles of the unmanned vehicle are shown in Fig. 17(b). From this figure, the unmanned vehicle successfully plans a safe obstacle avoidance path and makes a turn for obstacle avoidance at 4 s. The three-axis attitude angular acceleration and three-axis acceleration of the unmanned vehicle during travel are shown in Fig. 17(c). According to Table VI, the maximum yaw angular acceleration of the unmanned vehicle during obstacle avoidance is -12.1 rad/s$^2$, the maximum roll angular acceleration is 11.8 rad/s$^2$, and the maximum lateral acceleration is 0.56 g. The results show that the stability index in the obstacle avoidance process is less than the critical value of instability. Based on our proposed method, the obstacle avoidance process of unmanned vehicle is more efficient and stable.

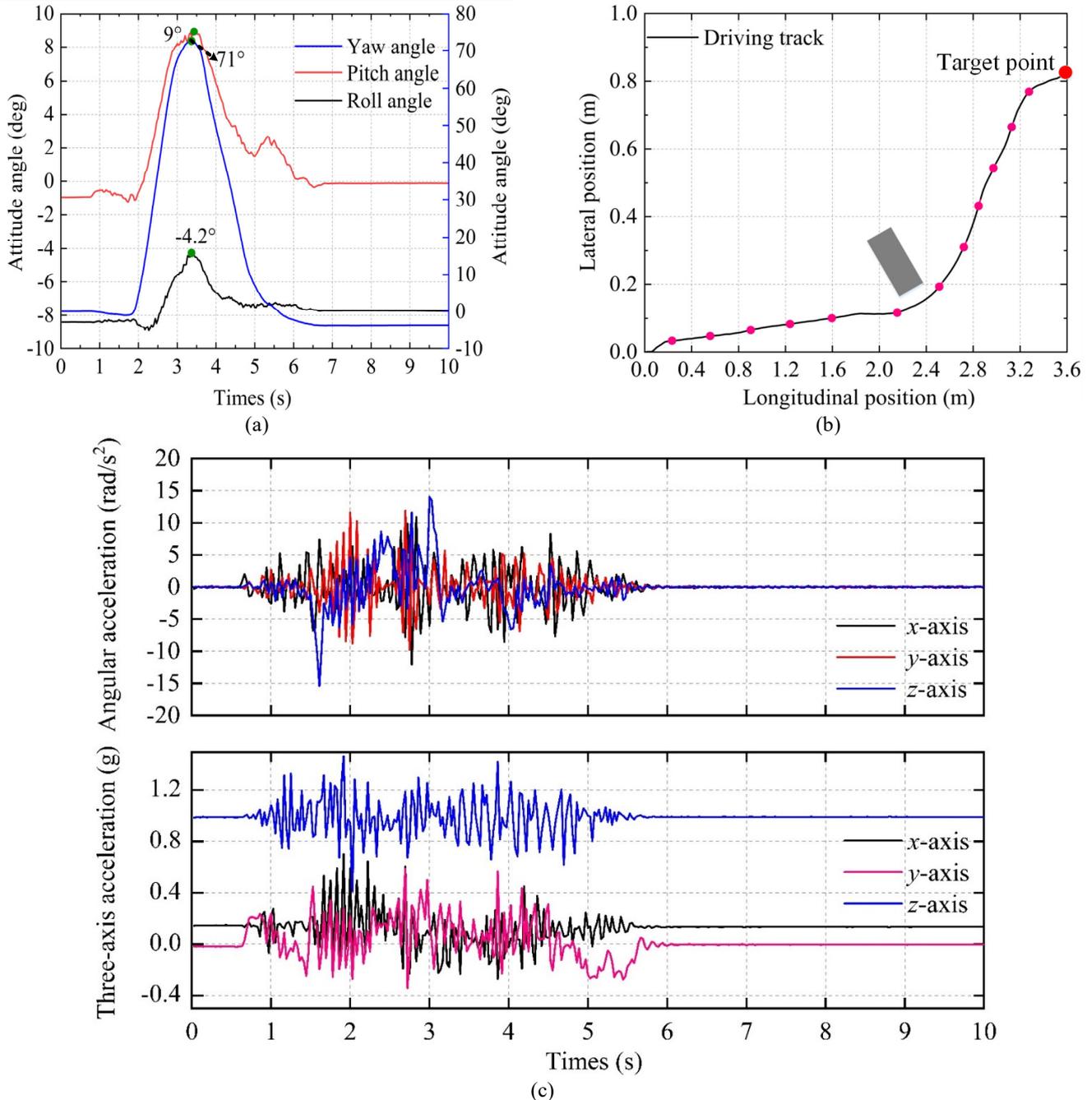

**Fig. 17.** Experimental results of Scenario 3 (longitudinal slope condition).

As shown in Table VI, the maximum DSE values of the three large-slope ground experiments are less than the critical value of instability required by the project, and no instability occurs in unmanned vehicles.

According to engineering requirements, we set the initial speed of the unmanned vehicle to 0.9 m/s. We conducted 72 vehicle obstacle avoidance experiments, and the success rate of the experiment was 98.7%. Note that our outdoor environment is conducted in rainy, snowy, and cold weather with relatively poor road conditions, which is very different from the training environment, bringing more difficulties to the deep learning algorithm and stable obstacle avoidance. In addition, it demonstrated the generalization ability and feasibility of our proposed model for unknown environments.

TABLE VI
TRIAXIAL ANGULAR ACCELERATION AND ACCELERATION OF VEHICLE

| Evaluating indicator | Experiment 1 | Experiment 2 | Experiment 3 |
|---|---|---|---|
| Slope angle (angle) | 9 | -7 | -6 |
| Around the x-axis (rad/s$^2$) | 7.76 | -12.13 | -12.10 |
| Around the y-axis (rad/s$^2$) | 6.61 | 8.11 | 11.87 |
| Around the z axis (rad/s$^2$) | 7.98 | -12.32 | -15.38 |
| x-axis (g) | 0.50 | 0.47 | 0.70 |
| y-axis (g) | 0.47 | -0.39 | 0.59 |
| z-axis (g) | 1.60 | 1.32 | 1.47 |
| DSE (max) | 3.11 | 4.29 | 5.63 |

## VI. CONCLUSION

This study presents a stable obstacle avoidance control method of tracked intelligent transportation vehicle in non-structural environment based on deep learning. The field site environment was simulated according to the configuration of the virtual simulation and the effectiveness of the method was verified through ground experiments.

The control method proposed in this study mainly consists of three parts: (1) obstacle information detection system, (2) expert obstacle avoidance strategy, and (3) Attention-LSTM. First, the system identifies and classifies obstacles in the field based on obstacle features and spatial relative position relationships to improve the usability of the data. Second, we formulated corresponding expert obstacle avoidance strategies based on different environmental obstacles. Finally, the deep learning model of Attention-LSTM with 12 inputs and 2 outputs was established to adjust the vehicle driving strategy in the current state. Moreover, we obtained a large number of data through joint simulation to train and test the model, and designed the ground experiment to verify the method proposed in this study. The method has shown good performance in terms of accuracy and reliability through simulations and ground experimental tests.

The two main contributions of our study are as follows:

(1) This study designed a vehicle obstacle avoidance control strategy combined with deep learning, data processing and vehicle instability mechanism in the non-structural environment. This can be used to solve path planning problems related to obstacle avoidance work in other similar engineering fields, such as multi-vehicle formation task assignment strategies for automated cooperative driving and multi-lane formation methods for unmanned vehicles.

(2) The control method proposed realized real-time obstacle avoidance planning for non-structural roads. If the method is transplanted to more complex field environments, unmanned vehicles can detect obstacles in advance and plan the best path to accomplish stable obstacle avoidance. In addition, the method proposed laid the foundation for future research on autonomous transport operations in the field.

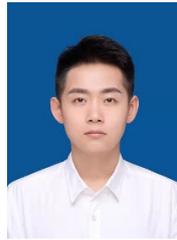

**Yitian Wang** received his B.S degrees in Mechanical Design, Manufacturing and Automation from Changchun University of Science and Technology in Changchun, China in 2016. He is currently pursuing a Ph.D. in the College of Instrumentation and Electrical Engineering, Jilin University. His research interests include path planning and pose estimation.

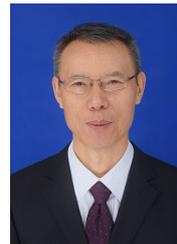

**Jun Lin** is an Academician of the Chinese Academy of Engineering, Beijing, China. He serves as the Director of the National Geophysical Exploration Equipment Engineering Research Center, Jilin University, Changchun, China. He actively conducts research in the geophysical exploration theory, technologies, and equipment.

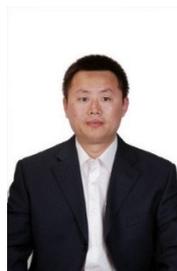

**Liu Zhang** was born in Bengbu, Anhui, China, in 1978. He received the Ph.D. degree from the Harbin Institute of Technology, in 2007. He is currently a Professor with the College of Instrumentation and Electrical Engineering, Jilin University. His research interests include aerospace optical remote sensing system design, simulation and application technology, and star sensor technology.

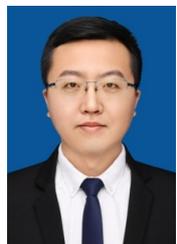

**Tianhao Wang** received the B.S. degree in electrical engineering and the Ph.D. degree in vehicle engineering from Jilin University, Changchun, Jilin, China, in 2010 and 2016, respectively. From 2016 to 2019, he was a Postdoctoral Researcher with the Department of Science and Technology of Instrument, Jilin University. He is currently an Assistant Professor with the College of Instrumentation and Electrical Engineering, Jilin University. His research interest includes numerical and experimental studies of crosstalk in complex cable bundles, with a particular emphasis on considering parameter variability using efficient statistical approaches.

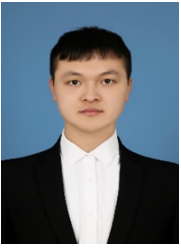

**Hao Xu** received his B.S. degree from Changchun University of Technology and the M.S. degree from Jilin University. He is currently pursuing the Ph.D. degree in the School of Instrument Science and Electrical Engineering, Jilin University. His research interests include path planning of intelligent robots.

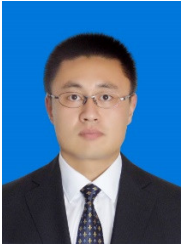

**Guanyu Zhang*** received the M.S. and Ph.D. degrees in mechanical engineering from Jilin University, in 2015. He is an associate professor in the Department of Instrument Science and Electrical Engineering, Jilin University. His research interests include mechatronics system design and artificial intelligence control technology research. (Corresponding author)

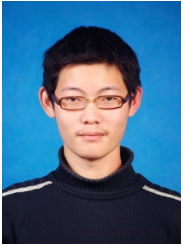

**Yang Liu*** received the Ph.D. degree from Jilin University in 2020 and is working in Jilin University as an assistant research fellow. His research interests include stereo vision, 3D reconstruction, texture filtering, and SLAM. (Corresponding author)